%% file: main.tex
\documentclass{article}

\usepackage{microtype}
\usepackage{graphicx}
\usepackage{subfigure}
\usepackage{booktabs} 

\usepackage{hyperref}
\usepackage{tcolorbox}

\usepackage[accepted]{icml2024}

\usepackage{amsmath}
\usepackage{amssymb}
\usepackage{mathtools}
\usepackage{amsthm}
\usepackage{enumitem}
\usepackage[capitalize,noabbrev]{cleveref}

\theoremstyle{plain}

\theoremstyle{definition}

\theoremstyle{remark}

\usepackage[textsize=tiny]{todonotes}

\icmltitlerunning{SpeechGPT-Gen: Scaling Chain-of-Information Speech Generation}

\usepackage{caption}
\newcommand{\blfootnote}[1]{%
  \begingroup
  \renewcommand\thefootnote{}\footnote{#1}%
  \addtocounter{footnote}{-1}%
  \endgroup
}

\begin{document}

\twocolumn[
\icmltitle{SpeechGPT-Gen: Scaling Chain-of-Information Speech Generation }



\icmlsetsymbol{equal}{*}
\icmlsetsymbol{correpsonding}{$\dagger $}

\begin{icmlauthorlist}
\icmlauthor{Dong Zhang}{equal}
\icmlauthor{Xin Zhang}{equal}
\icmlauthor{Jun Zhan}{}
\icmlauthor{Shimin Li}{}
\icmlauthor{Yaqian Zhou}{correpsonding}
\icmlauthor{Xipeng Qiu}{correpsonding}

\end{icmlauthorlist}


\center{School of Computer Science, Fudan University}

{\tt \{dongzhang22,xin\_zhang22\}@m.fudan.edu.cn}, 
{\tt 	\{zhouyaqian,xpqiu\}@fudan.edu.cn} \\

\url{https://0nutation.github.io/SpeechGPT-Gen.github.io/}

\icmlkeywords{Machine Learning, ICML}

\vskip 0.3in
]




\begin{abstract}
\input{Sections/000_abstract.tex}
\end{abstract}
\blfootnote{$^\ast$Equal contribution $^\dagger$Corresponding author}

\section{Introduction}
\input{Sections/010_intro.tex}

\section{Related Work and Background}
\input{Sections/020_related.tex}


\section{SpeechGPT-Gen}
\input{Sections/040_method}

\section{Experiments}
\input{Sections/050_exps}

\section{Analysis}
\input{Sections/060_analysis}

\section{Conclusion}
\input{Sections/070_conclusion.tex}



\bibliography{custom}
\bibliographystyle{icml2024}

\newpage
\appendix
\onecolumn

\input{Sections/080_appendix.tex}

\end{document}

%% file: Sections/000_abstract.tex
Benefiting from effective speech modeling, current Speech Large Language Models (SLLMs) have demonstrated exceptional capabilities in in-context speech generation and efficient generalization to unseen speakers. 
However, the prevailing information modeling process is encumbered by certain redundancies, leading to inefficiencies in speech generation.
We propose Chain-of-Information Generation (CoIG), a method for decoupling semantic and perceptual information in large-scale speech generation. Building on this, we develop SpeechGPT-Gen, an 8-billion-parameter SLLM efficient in semantic and perceptual information modeling. It comprises an autoregressive model based on LLM for semantic information modeling and a non-autoregressive model employing flow matching for perceptual information modeling. Additionally, we introduce the novel approach of infusing semantic information into the prior distribution to enhance the efficiency of flow matching. 
Extensive experimental results demonstrate that SpeechGPT-Gen markedly excels in zero-shot text-to-speech, zero-shot voice conversion, and speech-to-speech dialogue, underscoring CoIG's remarkable proficiency in capturing and modeling speech's semantic and perceptual dimensions.
Code and models are available at ~\url{https://github.com/0nutation/SpeechGPT}.

%% file: Sections/010_intro.tex

Large Language Models (LLMs) exemplified by ChatGPT~\cite{openai2023gpt4} and LLaMA~\cite{touvron2023llama} have demonstrated remarkable capabilities with expansive significant parameters. Scaling laws~\cite{kaplan2020scaling} indicate the model and data magnitude play a pivotal role in enhancing the performance.
For speech generative models, recent efforts have been made to scale up training data to nearly 100k hours of in-the-wild speech, which showcase significant capabilities in in-context speech generation, particularly in terms of generalization to unseen speakers.
~\cite{borsos2022audiolm, wang2023neural, le2023voicebox, shen2023naturalspeech, yang2023uniaudio, zhang2023speechtokenizer}. 
From an information modeling perspective, the current approach to large-scale speech generation can be classified into two types: 1) \textit{Integrated Generation}, which involves modeling both semantic and perceptual information simultaneously, without disentanglement during the speech generation process~\cite{wang2023neural,yang2023uniaudio}. 2) \textit{Semantic-Disentangled Generation}, which initially focuses on semantic modeling, followed by the integrated modeling of both semantic and perceptual information to generate complete speech~\cite{borsos2022audiolm,borsos2023soundstorm}.
However, although they achieved high-quality speech generation, inevitable redundancies exist in their modeling process, which exhibits potential inefficiency and necessitates complicated strategies.


\begin{figure}[t]
\centering
\includegraphics[width=1\columnwidth]{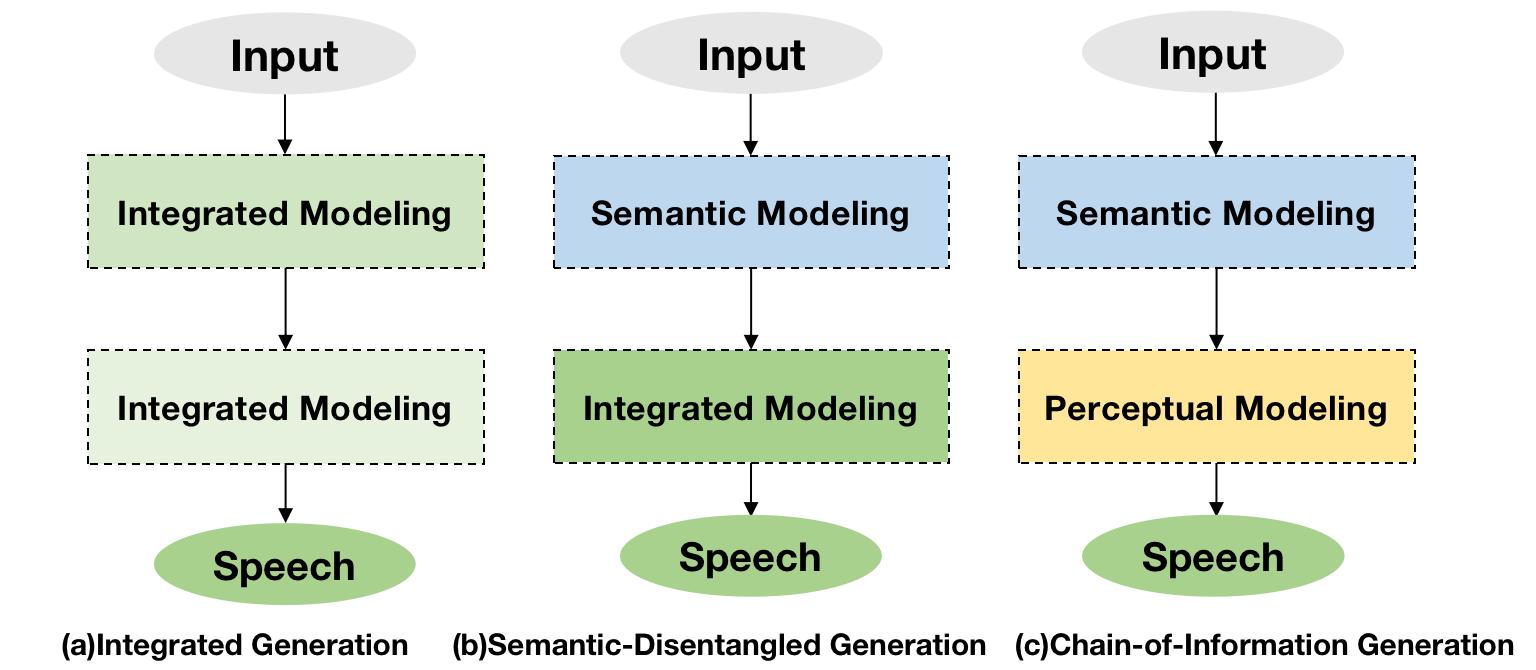} 
\caption{Illustration of three speech generation methods. Integrated modeling refers conductinguct semantic and perceptual modeling simultaneously. (a)~Integrated Generation (b)~Semantic-Disentangled Generation (c)~Chain-of-Information Generation}
\vspace{-0.5cm}
\label{fig:coi}
\end{figure}


Therefore, this paper presents \textit{Chain-of-Information Generation} (GoIG), a semantic and perceptual disentangled approach to large-scale speech generation.
Similar to the Chain-of-Thought~\cite{wei2023chainofthought}, Chain-of-Information decomposes the complex speech generation process into multiple intermediate steps, starting with the semantic modeling process and subsequently incorporating perceptual modeling to generate complete speech. 
Based on Chain-of-Information generation, we build SpeechGPT-Gen, a Speech Large Language Model (SLLM) with strong semantic and perceptual modeling abilities.
In terms of modeling, we employ SpeechTokenizer~\cite{zhang2023speechtokenizer} to extract two separate representations containing semantic and perceptual information, respectively. SpeechGPT-Gen consists of an LLM-based autoregressive model for semantic modeling and a flow-matching-based non-autoregressive model for perceptual modeling.  
Flow matching effectively models the transformation from a simple prior distribution to a complex data distribution and achieves promising results in speech generation~\cite{le2023voicebox, mehta2024matchatts}. We propose two modes of flow matching for perceptual modeling: 1) Explicit Chain, which directly models the probability path from a standard gaussian prior to the perceptual representation of speech. 2) Implicit Chain, which
models the probability path from a semantic-injected prior to the representation containing complete speech information, implicitly conducting perceptual modeling by capitalizing on the complementary relationship between semantic and perceptual information.

By scaling up the model parameters to 8 billion, SpeechGPT-Gen demonstrates strong semantic and perceptual modeling capabilities due to the efficiency of CoIG and achieves impressive performance on zero-shot text-to-speech, zero-shot voice conversion, and speech-to-speech dialogue.

Our contributions include the following:
 \begin{itemize}[itemsep=1pt, leftmargin=10pt, parsep=0pt, topsep=1pt]
    \item 
    We introduce Chain-of-Information Generation, a semantic and perceptual disentangled modeling approach to large-scale speech generation.

    \item
    We present SpeechGPT-Gen, a large speech langauge model with strong semantic and perceptual modeling abilities.

    \item 
    We propose to improve the efficiency of flow matching by injecting semantic information into its prior distribution.

    \item 
    We scale up the parameters of the speech generative model to 8 billion, achieving remarkable performance in zero-shot text-to-speech, zero-shot voice conversion, and speech-to-speech dialogue.

\end{itemize}

%% file: Sections/020_related.tex
\noindent\textbf{Large Scale Speech Generation}~
Recently, a series of speech generation works has focused on scaling up the amount of speech data to improve the generalization abilities. The first type of approach involves autoregressive modeling based on speech discrete representations, like VALL-E~\cite{wang2023neural} and USLM~\cite{zhang2023speechtokenizer}. They view the speech generation task as a conditional language modeling problem.
The second type utilizes non-autoregressive modeling based on diffusion or flow matching, such as NaturalSpeech2~\cite{shen2023naturalspeech} and Voicebox~\cite{le2023voicebox}. They learn to transform the simple prior distribution into a complex speech representation distribution. 
Both methods demonstrate strong generalization capabilities in speech generation tasks. However, from the information modeling perspective, certain redundancies exist in their modeling process.
SpeechGPT-Gen employs Chain-of-Information modeling approach and performs semantic and perceptual modeling sequentially.

\noindent\textbf{Disentangled Speech Modeling}~
Speech disentanglement is widely applied to traditional small-scale speech generation tasks like voice conversion~\cite{qian2020unsupervised} and speech encoding~\cite{polyak2021speech}, but not for large-scale speech generation yet.
SpeechTokenizer~\cite{zhang2023speechtokenizer} employs semantic surpervision on Residual Vector Quantization~(RVQ)-GAN to hierarchically disentangle different aspects of speech information across different RVQ layers. Specifically, the first RVQ quantizer generates tokens containing content information, while subsequent quantizers complement the remaining paralinguistic information.
We leverage the disentangled speech representations extracted by SpeechTokenizer to construct a large-scale speech generation model.


\noindent\textbf{Flow Matching for Speech Generation}~
Continuous Normalizing Flows~(CNFs)~\cite{chen2019neural} aims to estimate unknown distribution \( q(x) \) of data \( x \in \mathbb{R}^d \) by learning the probability path from a simple prior distribution \( p_0 \) to a data distribution \( p_1 \approx q \). 
CNFs defines the path through a time-dependent probability density function \( p_t : [0, 1] \times \mathbb{R}^d \rightarrow \mathbb{R}^+ \). The flow \( \phi_t : [0, 1] \times \mathbb{R}^d \rightarrow \mathbb{R}^d \), constructed by time-dependent vector field \( v_t : [0, 1] \times \mathbb{R}^d \rightarrow \mathbb{R}^d \), is defined via an ordinary differential equation (ODE):
\begin{align}
\frac{d}{dt}\phi_t(x) = v_t(\phi_t(x)); \quad \phi_0(x) = x 
\end{align}
So the time-dependent probability density function \( p_t \) is derived through the change of variables formula:
$ p_t = p_0(\phi_{t}^{-1}(x)) \det \left(\frac{\partial\phi_{t}^{-1}}{\partial x}(x)\right) $. The Flow Matching objective can be formulated as:
\begin{align}
L_{\text{FM}}(\theta) = \mathbb{E}_{t \sim U[0, 1],x \sim p_t(x)} \| u_t(x) - v_t(x; \theta) \|^2 \end{align}
where \( u_t \) is a vector field that generates probability path \( p_t \) and \( v_t(x; \theta) \) is the vector field parameterized by a neural network \( \theta \).
However, $L_{\text{FM}}$ is uncomputable for lack of prior knowledge of \( p_t \) or \( v_t \).
Luckily, \cite{lipman2023flow} proposed Conditional Flow Matching (CFM) objective presented as:
\begin{align}
L_{\text{CFM}}(\theta) = \mathbb{E}_{t,q(x_1),p_t(x|x_1)} \| u_t(x | x_1) - v_t(x; \theta) \|^2. 
\end{align}
by conditioning \( p_t \) and \( v_t \) on real data $x_1$ and prove FM and CFM have identical gradients w.r.t. $\theta$ for training the generative model.
We can adopt the optimal transport (OT) path as the conditional flow~\cite{lipman2023flow}, defined as \( p_t(x | x_1) = \mathcal{N}(x | t x_1, (1 - (1 - \sigma_{\text{min}})t)^2 I) \) and \( u_t(x | x_1) = (x_1 - (1 - \sigma_{\text{min}})x) / (1 - (1 - \sigma_{\text{min}})t) \) with a sufficiently small \( \sigma_{\text{min}} \) such that \( p_1(x | x_1) \) is centered around \( x_1 \).

Utilizing $L_{\text{CFM}}$ as the training objective function, some studies have successfully employed flow matching for speech generation, demonstrating promising results~\cite{le2023voicebox, liu2023generative, vyas2023audiobox, mehta2024matchatts}. However, they typically model the process from a simple prior distribution to mel-spectrograms~\cite{le2023voicebox} or latent features~\cite{vyas2023audiobox} that contains complete speech information. In contrast, SpeechGPT-Gen employs flow matching to model the perceptual information in speech and enhances its efficiency by modifying the prior distribution.

%% file: Sections/040_method.tex
Building upon the Chain-of-Information Generation method, SpeechGPT-Gen conducts semantic modeling and perceptual modeling sequentially. SpeechTokenizer~\cite{zhang2023speechtokenizer} is utilized to extract semantic representations and perceptual representations.
SpeechGPT-Gen consists of a LLM-based autoregressive model for semantic modeling and a flow matching-based non-autoregressive model for perceptual modeling.

\begin{figure*}[t] 
    \setlength{\abovecaptionskip}{-0.cm}
    \centering 
    \includegraphics[width=0.9\textwidth]{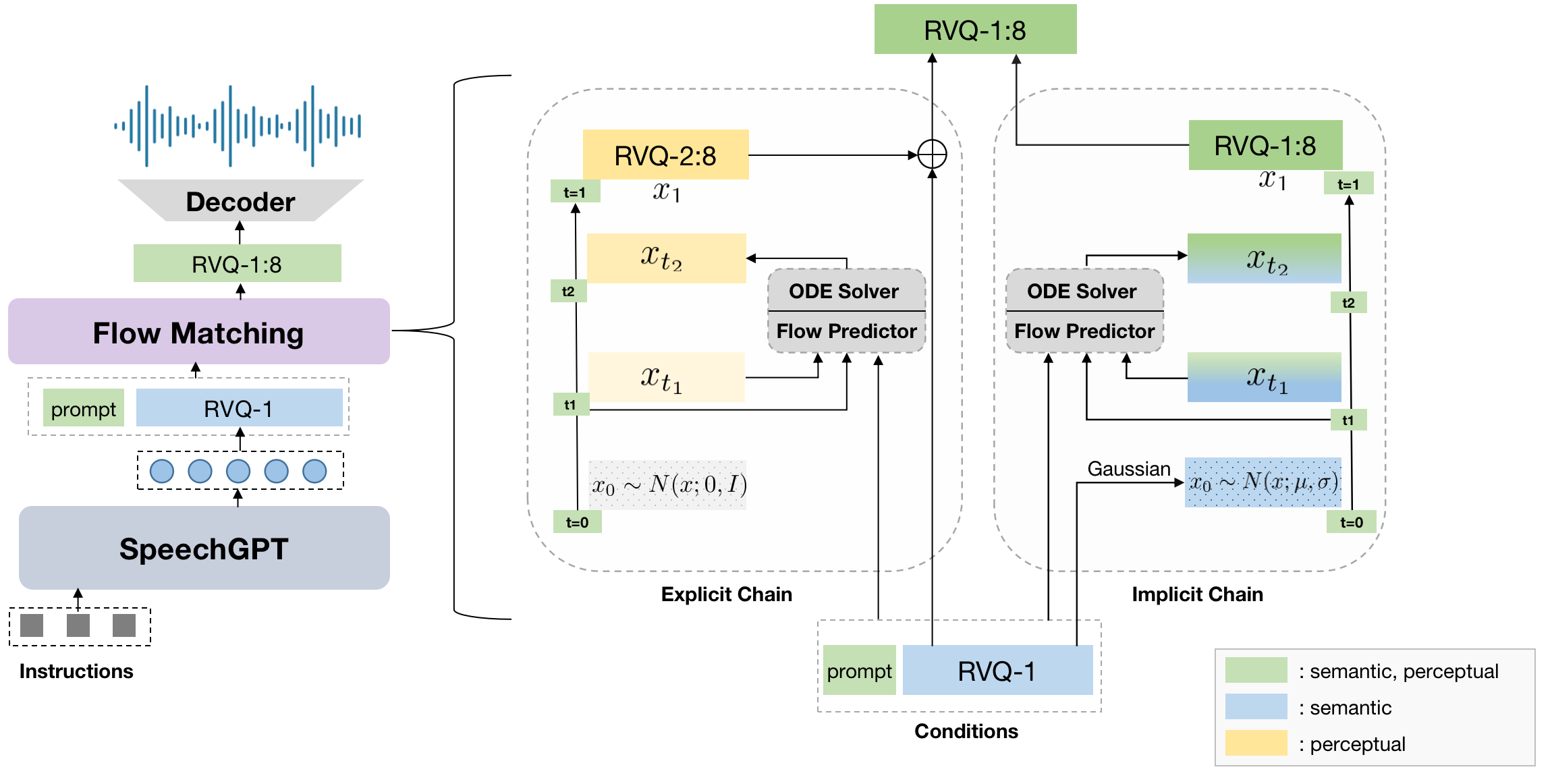} 
    \caption{SpeechGPT-Gen Overview. Decoder refers to SpeechTokenizer decoder. Blocks with different colors stand for representations containing different information.}
    \label{fig:model} 
\end{figure*}

\subsection{Speech Tokenization}
\label{sec:041_tokenizer}
\input{Sections/041_tokenizer}

\subsection{LLM for Semantic Modeling}
\label{sec:042_llm}
\input{Sections/042_llm}

\subsection{Flow Matching for Perceptual Modeling}
\label{sec:043_cnf}
\input{Sections/043_cnf}

\subsection{Application}
\label{sec:044_application}
\input{Sections/044_application}

%% file: Sections/041_tokenizer.tex
SpeechTokenizer~\cite{zhang2023speechtokenizer} is a Residual Vector Quantization~(RVQ)-based speech tokenization method and hierarchically disentangles different aspects of speech information across different RVQ layers. Specifically, the first RVQ quantizer generates tokens containing semantic information, while subsequent quantizers complement the remaining perceptual information.
SpeechTokenizer takes a single-channel audio signal denoted
as \(s \in \mathbb{R}^{d \cdot f_{sr}}\) as input, where \(d\) denotes duration and \(f_{sr}\) denotes the sampling rate.
The output of SpeechTokenizer comprises \(Q = 8\) hierarchical RVQ tokens \((q_1, \ldots, q_Q)\) with $Q$ codebooks \((C_1, \ldots, C_Q)\) for corresponding RVQ layer. $q_i \in \mathbb{R}^{T}$ represents a one-dimensional sequence with a length of $T$. 
RVQ tokens $q_i$ can be embed by codebook \(C_i \in \mathbb{R}^{K \times H}\), resulting in continuous vector sequence \(v_i \in \mathbb{R}^{T \times H}\), where \(v_i^j = C_i(q_i^j)\) for \(i \in 1, \ldots, Q\), \(j \in 1, \ldots, T\).
The sum of continuous representations from all RVQ layers denoted as $v_{1:8} = \sum_{i=1}^{Q}v_i$, encompasses all the information within the speech.
We utilize tokens from the first RVQ layer $q_1$ as representations containing semantic information, and continuous vectors summed from the second to the last layers $v_{2:7} = \sum_{i=2}^{Q}v_i$ as representations containing perceptual information. We train SpeechTokenizer on CommonVoice~\cite{ardila2020common}, and LibriSpeech~\cite{librispeech} dataset and other settings follow the same as ~\cite{zhang2023speechtokenizer}.

%% file: Sections/042_llm.tex
SpeechGPT~\cite{zhang-etal-2023-speechgpt} is a large cross-modal language model that can perform cross-modal instruction following and speech-to-speech dialogue, showing excellent speech semantic modeling capabilities.
Utilizing HuBERT~\cite{hsu2021hubert} to discretize speech into units, SpeechGPT performs modality-adaptation pretraining, cross-modal instruction fine-tuning, and chain-of-modality instruction fine-tuning. We employ SpeechGPT for our speech semantic modeling by replacing the HuBERT discrete units with SpeechTokenizer RVQ-1 tokens $q_1$.
For model structure, we adopt LLaMA2-7B-Chat as our pretrained LLM.
For training data, we adopt SpeechTokenizer RVQ-1 tokens to represent speech and follow the process and settings of SpeechInstruct~\cite{zhang-etal-2023-speechgpt} to construct the training dataset $D$, including cross-modal instruction set and chain-of-modality instruction set. We use Multilingual Librispeech~\cite{Pratap_2020}, Gigaspeech~\cite{chen2021gigaspeech}, Commonvoice~\cite{ardila2020common} and Librispeech~\cite{librispeech} to construct cross-modal instruction set.
For training, we skip the modality-adaptation pre-training and perform cross-modal instruction fine-tuning on LLaMA2-7B-Chat with cross-modal instruction set $I$. 
In chain-of-modality instruction fine-tuning, we choose full-parameter fine-tuning instead of LoRA fine-tuning on the chain-of-modality instruction set.
The training objective for both training stages can be formatted as:
\begin{align}
L_{ar} = -\log \prod_{t=0}^{n} P(d_{t} | d_{<t}; \theta_{ar})
\end{align}
where \(d \in D\) consisting of \(d_1, \ldots, d_{n}\), \(n\) is the total number of tokens in sample \(d\) and $\theta_{ar}$ denotes the parameters of SpeechGPT.

%% file: Sections/043_cnf.tex

Flow matching effectively models the transformation from a simple prior distribution to a complex
data distribution and has achieved promising results in
speech generation. We perform perceptual modeling with flow-matching.
As shown in Figure~\ref{fig:model}, given speech $s$, semantic representation $v_1$, perceptual representation $v_{2:8}$ and the complete information representation $v_{1:8}=v_1+v_{2:8}$ extracted by SpeechTokenizer, perceptual modeling refers to predicting the complete representation $v_{1:8}$ given the prompt speech $a$ and the semantic representation $v_1$. 
We propose two modes of flow matching for perceptual modeling: Explicit Chain and Implicit Chain.

\noindent\textbf{Explicit Chain}~
For explicit chain, the model directly generates perceptual representations from a standard gaussian prior, conditioned on semantic representations and speech prompts.
Concretely, flow matching models the paths that transform from a simple standard gaussian prior distribution \(p_0= N(x_0 | \mu = 0, \sigma = 1)\) to the data distribution \(p_1\) of perceptual representation $v_{2:8}$ in a continuous manner. We use flow step \(t\) to describe the progress of transformation, where the prior is at \(t = 0\) and the data is at \(t = 1\).
We let $x_1=v_{2:8}$, prior $p_0 = N(x_0 | \mu = 0, \sigma = 1)$ and $z=v_1$.

\noindent\textbf{Implicit Chain}~
For implicit chain, the model does not explicitly generate perceptual representation. Instead, it leverages the complementary relationship between semantic representation \(v_1\) and perceptual representation $v_{2:8}$, that is $v_{2:8}=v_{1:8}-v_1$. 
Concretely, we inject semantic information into prior distribution to get \textit{semantic prior}: 
\begin{align*}
p_0 = N(x_0 | \mu = v_1, \sigma = 1)
\end{align*}
By modeling the paths that transform from semantic prior \(p_0\) to the data distribution \(p_1\) of complete information representation $v_{1:8}$ in a continuous manner, the model learns to supplement perceptual information for semantic information to obtain the complete information, implicitly performs perceptual modeling.
We use flow step \(t\) to describe the progress of transformation, where the prior is at \(t = 0\) and the data is at \(t = 1\).
We set $x_1=v_{1:8}$, prior $p_0 = N(x_0 | \mu = v_1, \sigma = 1)$ and $z=v_1$.

\noindent\textbf{Training}~
After determining \(x_1\), \(p_0\), and \(z\), we train the perceptual modeling model with flow matching.
Let $m$ be a binary temporal mask which is of the same length $N$ as \(x_1\), and \(x_{\text{tgt}} = m \odot x_1\) and \(x_{\text{pmt}} = (1 - m) \odot x_1\) be the complementary masked versions of \(x_1\). The generative model learns \(p(x_{\text{tgt}} | z, x_{\text{pmt}})\) with \(z\) and \(x_{\text{pmt}}\) as the condition, and \(x_{\text{tgt}}\) as the predicted data.
$m[:n]=0$,$m[n:]=1$, where $n \sim \mu(0,N)$.
We use Conformer~\cite{gulati2020conformer} encoder parameterized by $\theta$ to predict the conditional vector field \(v_t(x_t, x_{\text{pmt}}, z; \theta)\), where \(x_t\) represents a sample at flow step \(t\). 
The sequences (\(x_t\), and \(z\)) are frame-wise concatenated and projected then concatenated with \(x_{\text{pmt}}\) before sending into Conformer as input.
During training, for $x$ and a prior sample \(x_0 \sim p_0\), we have
\begin{align}
x_t = (1 - (1 - \sigma_{\text{min}})t)x_0 + tx_1 \\ 
u_t(x_t | x_1) = x_1 - (1 - \sigma_{\text{min}})x_0
\end{align}
We set $\sigma_{\text{min}}=0$.
The training objective can be formulated as
\begin{align*}
L_{nar} &= \mathbb{E}_{t, m, q(x_1, z), p_0(x_0)} \\&\quad \left\|m \odot (u_t(x_t | x_1) - v_t(x_t, x_{\text{pmt}}, z; \theta_{nar}))\right\|^2,
\end{align*}
 Multilingual LibriSpeech is utilized for training, and the model structure is a standard Conformer network with bidirectional self-attention and rotary positional embeddings~\cite{su2023roformer}. 

\noindent\textbf{Inference}~
Given speech prompt $a$ and semantic representation $v_1$ derived from SpeechGPT, we first utilize SpeechTokenizer to extract complete information representation $v_{1:8}^a$ for speech prompt. For explicit chain, we set prior $p_0 = N(x_0 | \mu = 0, \sigma = 1)$. For implicit chain, we set $p_0 = N(x_0 | \mu = v_1, \sigma = 1)$.
We first sample $x_0$ from prior $p_0$ and then utilize an ODE solver to compute \(\phi_1(x_0)\) by evaluating \(v_t\) at multiple \(t\) to approximate the integration from \(t = 0\) to \(t = 1\) given the initial condition \(\phi_0(x_0) = x_0\) and \(\frac{d\phi_t(x_0)}{dt} =v_t(x_t, v_{1:8}^a, v_1; \theta)\). At each step, the ODE solver estimates \(x_{t+1}\) given \(t\), \(x_t\) from the previous step, the speech prompt $v_{1:8}^a$, semantic representation $v_1$ and the model $\theta$. After the final step, it produces \(x_1\). For the explicit chain, \(x_1\) contains perceptual information, and it needs to be added to \(v_1\) to obtain complete information representation \(\hat{v_{1:8}}=x_1+v_1\). For implicit chain, \(x_1\) contains the whole speech information, so we have \(\hat{v_{1:8}}=x_1\).  Finally, we use the SpeechTokenizer decoder to generate the speech output conditioned on $v_{1:8}$.

%% file: Sections/044_application.tex
\noindent\textbf{Zero-shot Text-to-Speech}~
SpeechGPT-Gen can synthesize speech with the timbre of a given prompt by taking the given text as input for SpeechGPT, concatenating the output with the prompt speech, utilizing the flowing matching model, and transforming the output into the synthesized speech.

\noindent\textbf{Zero-shot Voice Conversion}~
SpeechGPT-Gen can transform the timbre of a source speech to match a given prompt speech. It achieves this by using the prompt speech and SpeechTokenizer RVQ-1 tokens of the source speech as input for the flowing matching model. The resulting output is processed to generate the final converted speech.

\noindent\textbf{Speech-to-Speech Dialogue}~
SpeechGPT-Gen can respond to a speech instruction with speech that matches the timbre of a given speech prompt.
It achieves this by using the SpeechTokenizer RVQ-1 tokens of the speech instruction as input for SpeechGPT, concatenating the output with the prompt speech, utilizing the flowing matching model, and transforming the output into the synthesized speech.

%% file: Sections/050_exps.tex
\subsection{Setups}
\label{sec:051_setup}
\input{Sections/051_setups}

\subsection{Metrics}
\label{sec:052_metrics}
\input{Sections/052_metric}


\input{Tables/ttsvc}
\input{Tables/dialogue}

\subsection{Main Results}
\label{sec:054_main}
\input{Sections/054_main}

%% file: Sections/051_setups.tex
\textbf{Model}~
We adopt LLaMA2-7B-Chat~\cite{touvron2023llama} as pretrained model for SpeechGPT. Flow matching model consists of a Conformer encoder with the hidden size 1024, 8 attention heads, and 4096 FFN hidden states.  We implement the Base and Large versions of the flow matching model, with layers of 12 and 24, respectively.

\textbf{Training}~
We train SpeechGPT-Gen from LLaMA2-7B-Chat. For cross-modal instruction fine-tuning, we trained the model for 77000 steps with a batch size of 1152 and a maximum sequence length of 1024 on 24 A100 GPUs. For chain-of-modality instruction fine-tuning,
we train for 1000 steps with batch size 512 and maximum sequence length 1536 on 16 A100 GPUs. 
We list model configuration for the flow matching model in Appendix~\ref{sec:app:model_config}.

\textbf{Baselines}~
For zero-shot TTS task, we adopt YourTTS~\cite{casanova2023yourtts}, VALL-E~\cite{wang2023neural}, NaturalSpeech 2~\cite{shen2023naturalspeech}, Voicebox~\cite{le2023voicebox} and USLM~\cite{zhang-etal-2023-speechgpt} as our baselines. YourTTS is a flow-base model adapted from VITS~\cite{kim2021conditional}. VALL-E and USLM are built on the neural codec and consist of an autoregressive and a non-autoregressive model. NaturalSpeech 2 is built on a diffusion model, and Voicebox is built on a continuous normalization flow model.
For zero-shot voice conversion task, we adopt YourTTS~\cite{casanova2023yourtts}, Soundstorm~\cite{borsos2023soundstorm} and USLM~\cite{zhang2023speechtokenizer} as our baseline systems.
For the speech-to-speech dialogue task, we adopt SpeechGPT~\cite{zhang-etal-2023-speechgpt} as our baseline system.

%% file: Sections/052_metric.tex
\textbf{Word Error Rate}~
can be used to evaluate the content correctness of synthesized speech by calculating the distance between the synthesized speech’s transcription and the input text. We use Whisper medium-en model~\cite{radford2022robust} to transcribe the synthesized speech.

\textbf{Speaker Similarity}~
measures the timbre consistency between synthesized speech and prompt speech. This is measured by the similarity between the speaker embedding of generated speech and that of the speech prompt. We calculate the similarity
with the following steps: 1) we utilize speaker embedding extractor~\footnote{For zero-shot TTS, we use \url{https://github.com/microsoft/UniSpeech/tree/main/downstreams/speaker_verification}. For zero-shot voice conversion, we use \url{https://huggingface.co/microsoft/wavlm-base-plus-sv}.} to calculate the speaker embedding for the
generated speech and the prompt speech. 2) we calculate the cosine similarity between the normalized
embeddings.

\textbf{QMOS}~
reflects the quality and naturalness of speech. We engaged 6 native speakers as contributors for QMOS evaluations. QMOS spans from 1 to 5, with higher values signifying greater speech quality.

\textbf{SMOS}~
reflects the timbre similarity between speech prompts and
generated speech. We engaged 6 native speakers as contributors for SMOS evaluations. SMOS spans from 1 to 5, with higher values signifying greater speech similarity.

\textbf{ChatGPT Score}~
evaluates the response quality for speech-to-speech dialogue tasks. We leveraged the Whisper medium-en model to transform the speech into its corresponding text, which was subsequently submitted for evaluation. Then we feed the prompt in appendix~\ref{sec:app:prompt_chatgpt_score} to ChatGPT to score the model's outputs based on response quality, with scores ranging from 1 to 5.

%% file: Tables/ttsvc.tex
\begin{table*}[ht]
\setlength{\belowcaptionskip}{-0.2cm}
 \renewcommand{\arraystretch}{1.2}
 \Large
 \centering
 \resizebox{0.9\textwidth}{!}{
 \begin{tabular}{l|cccc|cccc}
 \toprule
 & \multicolumn{4}{c}{Text-to-Speech} & \multicolumn{4}{c}{Voice Conversion} \\
 Model & WER~($\downarrow$) & SIM~($\uparrow$) & QMOS~($\uparrow$)   & SMOS~($\uparrow$) & WER~($\downarrow$) & SIM~($\uparrow$)  & QMOS~($\uparrow$) & SMOS~($\uparrow$) \\
 \midrule
 Groundtruth & 2.2 & -  &  3.97 & - & 1.9 &  0.93  & 3.99 & -  \\
 \midrule
 \multicolumn{9}{l}{\textit{Baselines}} \\
 YourTTS~\cite{casanova2023yourtts} & 7.7 & 0.34 & 3.42  &  3.14 & 10.1 & 0.72 & 3.48 & 3.25  \\
 VALL-E~\cite{wang2023neural}  & 5.9 & 0.58 & -  &  - & -  & -  & -  & - \\
 NaturalSpeech 2~\cite{shen2023naturalspeech}  & 2.3 & 0.62   &  - & -  & -  & -  & -  & - \\
 Voicebox~\cite{le2023voicebox}  & \textbf{1.9} & \textbf{0.68} & -  & -  & -  & -  & -  & - \\
 SoundStorm~\cite{borsos2023soundstorm} &- & - & - & - &  7.7 &  0.81 & 3.59 & 3.41 \\
 USLM~\cite{zhang2023speechtokenizer} & 5.4 & 0.56 & 3.61 & 3.42  & 6.4 &  0.83  & 3.65 & 3.48  \\
 \midrule
 SpeechGPT-Gen~(Explicit Chain) & 3.1 & 0.63 & 3.63 & 3.48 & 4.8 & 0.84 & 3.61 & 3.50 \\
 SpeechGPT-Gen~(Implicit Chain) & 2.4  &  0.66  & \textbf{3.69} & \textbf{3.51} &  \textbf{3.1}  &  \textbf{0.86}  & \textbf{3.72} & \textbf{3.54} \\
 \bottomrule
 \end{tabular}}
 \caption{Results of zero-shot text-to-speech and voice conversion.}
 \label{tab:main}
\end{table*}

%% file: Tables/dialogue.tex
\begin{table*}[ht]
\setlength{\belowcaptionskip}{-0.2cm}
    \renewcommand{\arraystretch}{1.2}
    \Large
    \centering
    \resizebox{0.7\textwidth}{!}{
    \begin{tabular}{l|cccc}
    \toprule
    & ChatGPT Score~($\uparrow$) & SIM~($\uparrow$) & QMOS~($\uparrow$) & SMOS~($\uparrow$) \\
    \midrule
    Groundtruth & 3.93  &  -  &  3.97 & - \\
    \midrule
    SpeechGPT~\cite{zhang-etal-2023-speechgpt}  & \textbf{3.63}  & 0.42 & 3.59  &  2.15   \\
    SpeechGPT-Gen~(Explicit Chain) & 3.60 & 0.83 &  3.60  &  3.39  \\
    SpeechGPT-Gen~(Implicit Chain) &  3.61 & \textbf{0.87} & \textbf{3.62}  &  \textbf{3.44}  \\
    \bottomrule
    \end{tabular}}
    \caption{Results of speech-to-speech dialogue. }
    \label{tab:dialogue}
\end{table*}

%% file: Sections/054_main.tex
\textbf{Zero-shot TTS}~
Table~\ref{tab:main} presents the zero-shot TTS results of SpeechGPT-Gen and baseline systems. The test set is constructed from Librispeech test-clean, including 40 speakers. For each speaker, we randomly crop a 3-second-long sample as the prompt and the textual content of another 4 to 10-second-long sample as the input text. 
SpeechGPT-Gen with implicit chain achieves a lower WER than discrete token-based methods like VALL-E and USLM. It also achieves a WER close to non-autoregressive methods like NaturalSpeech 2 and Voicebox, indicating that the speech generated by SpeechGPT-Gen has relatively high content accuracy, demonstrating the powerful capability of LLM-based semantic modeling. SpeechGPT-Gen with implicit chain also achieves close speaker similarity compared to Voicebox, proving its significant advantage in maintaining coherent speech timbre. The optimal performance of SpeechGPT-Gen on QMOS and SMOS further demonstrates that the generated speech excels subjectively in speech quality and similarity compared to other baseline systems.
SpeechGPT-Gen with implicit chain outperforms SpeechGPT-Gen with explicit chain slightly across all metrics, indicating that it's more effective to conduct perceptual modeling implicitly with the same semantic modeling capability.

\textbf{Zero-shot Voice Conversion}~
Table~\ref{tab:main} presents the zero-shot voice conversion results of SpeechGPT-Gen and baseline systems. The test set is constructed from the VCTK dataset, which includes 109 speakers. For each speaker, a 3-second-long sample was randomly cropped as the prompt, and another 4 to 10-second-long sample served as the source speech.
YourTTS, SoundStorm, and USLM are utilized as our baseline models. We trained SoundStorm with HuBERT L9 units as semantic tokens and Encodec tokens as acoustic tokens on the Multilingual LibriSpeech dataset.
SpeechGPT-Gen with implicit chain achieves the lowest WER and the highest speaker similarity compared to all baseline models, showcasing the powerful in-context speech generation capability of flow matching based perceptual modeling. The optimal performance of SpeechGPT-Gen on QMOS and SMOS further demonstrates that the generated speech is better subjectively in both speech quality and speech similarity compared to other baseline systems.
SpeechGPT-Gen with implicit chain slightly outperforms SpeechGPT-Gen with explicit chain, indicating that utilizing flow matching for implicit perceptual modeling is more effective.

\textbf{Speech-to-Speech Dialogue}~
Table~\ref{tab:dialogue} presents the speech-to-speech dialogue results of SpeechGPT-Gen and the baseline system SpeechGPT.
For a fair comparison, we trained SpeechGPT using the LLaMA2-7B-CHAT as the base model, utilizing the same mHuBERT tokens as SpeechGPT~\cite{zhang-etal-2023-speechgpt}. The training process and settings were identical to those described in Section~\ref{sec:042_llm}.
We evaluate the speech-to-speech instruction-following task proposed in \cite{zhang-etal-2023-speechgpt}. The processing progress, test dataset, and evaluation metric align with those described in \cite{zhang-etal-2023-speechgpt}, where CHATGPT Score is used as the metric.
For semantic quality, SpeechGPT-Gen achieves a ChatGPT Score close to SpeechGPT, indicating a similar capability in speech understanding and semantic modeling. It indicates that SpeechGPT-Gen can correctly understand spoken questions and generate responses with appropriate semantics.
For perceptual quality, speech responses generated by SpeechGPT-Gen exhibit a significant advantage in speaker similarity compared to SpeechGPT. This is because SpeechGPT can only generate speech with a fixed voice. The observed improvement in QMOS also attests to SpeechGPT-Gen's ability to produce higher-quality speech.

%% file: Sections/060_analysis.tex
\subsection{CoIG is Effective and Efficient}
\label{sec:coi}
\input{Sections/061_coi_analysis}

\subsection{Semantic Prior Improves Flowing Matching}
\label{sec:prior}

\input{Sections/062_prior_analysis}

\subsection{Scalability}
\label{sec:scaling}
\input{Sections/063_scaling_analysis}

\subsection{Continuous Modeling vs. Discrete Modeling}
\label{sec:cont_vs_dist}
\input{Sections/064_cont_dist_analysis}

%% file: Sections/061_coi_analysis.tex
\begin{figure}[t]
    \centering
    \includegraphics[width=\linewidth]{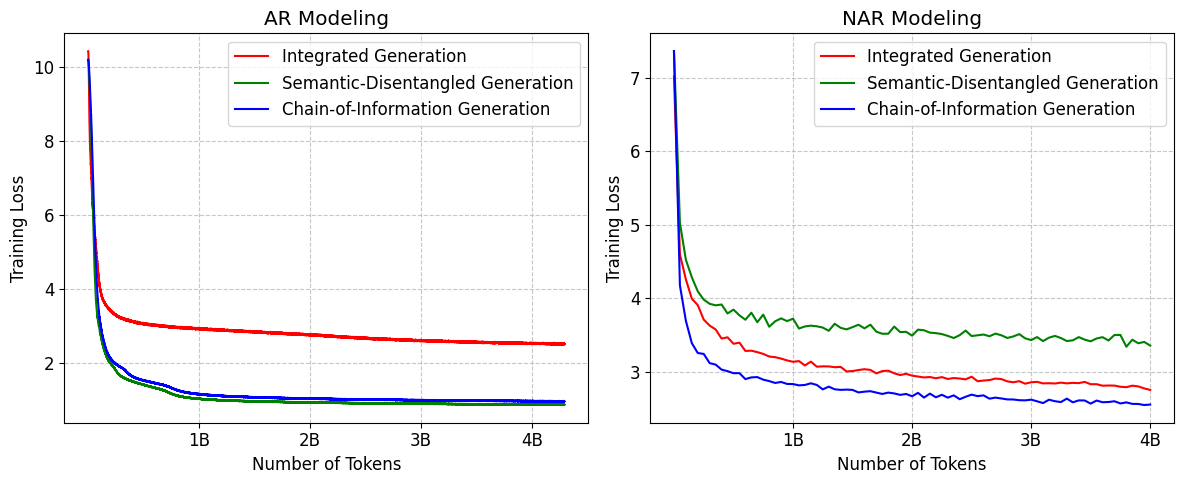}
    \caption{Training loss of AR modeling~(\textbf{Left}) and NAR modeling~(\textbf{Right}) for Integrated Generation, Semantic-Disentangled Generation and Chain-of-Information Generation.}
    \label{fig:coi_training}
\vspace{-0.4cm}
\end{figure}

\begin{figure}[t]
    \centering
    \includegraphics[width=\linewidth]{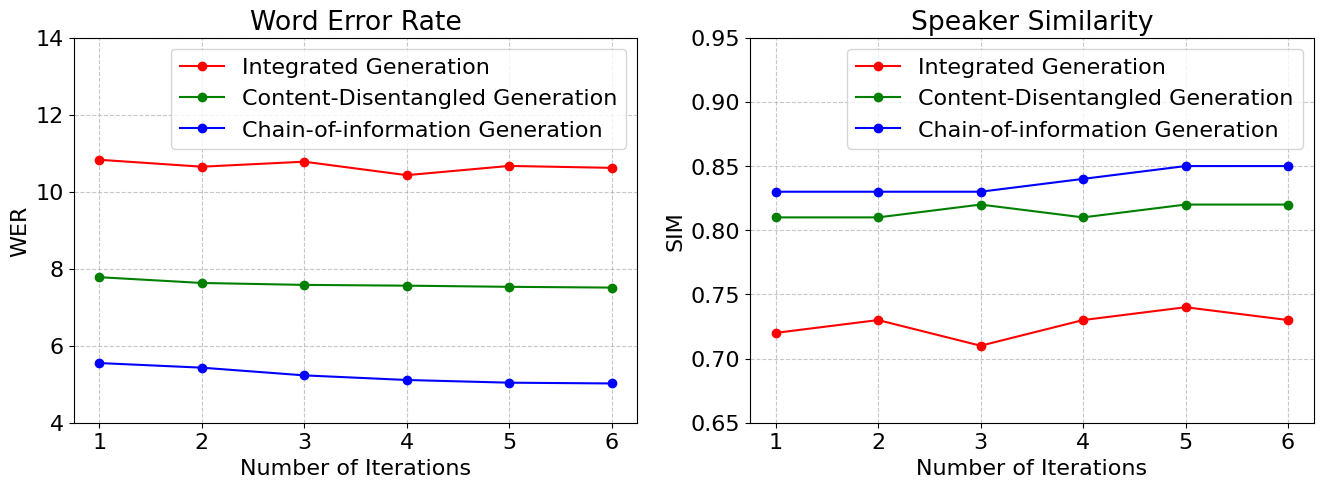}
    \caption{WER~(\textbf{Left}) and speaker similarity~(\textbf{Right}) of zero-shot TTS for Integrated Generation, Semantic-Disentangled Generation and Chain-of-Information Generation.}
    \vspace{-0.4cm}
    \label{fig:coi_infer}
\end{figure}

We conduct experiments to verify whether Chain-of-Information modeling is more effective and efficient. We compared the model training convergence and performance on zero-shot TTS among three speech generation modeling methods: Integrated Generation~(IG), Semantic-Disentangled Generation~(SDG), and Chain-of-Information Generation~(COIG).
For a fair comparison, we configure the training data, model structure, and hyperparameters to be identical. They all employ a two-stage modeling approach consisting of an autoregressive (AR) model and a non-autoregressive (NAR) model. The AR model utilizes a transformer decoder-only architecture, while the NAR model adopts a Conformer architecture with the SoundStorm procedure. We list the model configuration in Appendix~\ref{sec:app:model_config}. All models are trained on the Multilingual LibriSpeech dataset.
\textbf{Integrated Generation} generates semantic and perceptual information simultaneously. We choose EnCodec to extract discrete speech representations. The AR model learns the sequence-to-sequence mapping from the input raw text to the tokens of the first RVQ layer in EnCodec. The NAR model is trained to generate EnCodec tokens from the second to last RVQ layers, conditioned on the first RVQ layer tokens and the speech prompt.
\textbf{Semantic-Disentangled Generation} initially focuses on semantic modeling, followed by the integrated modeling of both semantic and perceptual information. We adopt HuBERT~\cite{hsu2021hubert} to extract semantic tokens and EnCodec to extract acoustic tokens. The AR model performs semantic modeling by converting the input text to semantic tokens as a casual language modeling task. Conditioned on the semantic tokens and the speech prompt, the NAR model performs integrated modeling by generating EnCodec tokens of all RVQ layers containing all speech information.
\textbf{Chain-of-Information Generation} conducts semantic modeling first, followed by perceptual modeling to generate complete speech information. We use SpeechTokenizer to extract speech tokens. The AR model learns to transform the input text to the tokens from the first RVQ layer of SpeechTokenizer. The NAR model is trained to generate the second to last RVQ layer tokens conditioned on the first RVQ layer tokens and the speech prompt.

\noindent\textbf{Model Convergence}~
We first inspect the training loss of AR modeling and NAR modeling for different generation methods, as presented in Figure~\ref{fig:coi_training}. For AR modeling, the training loss decrease speed of SDG and COIG is noticeably faster than that of IG. For NAR modeling, COI achieves the fastest convergence compared to IG and SDG, resulting in improved training efficiency. SDG exhibits the highest final NAR training loss because its NAR model needs to generate complete speech information, posing the greatest difficulty.

\noindent\textbf{Downstream Performance}~
We evaluate the performance of three-generation methods on zero-shot TTS. The test set is constructed from the VCTK dataset. For each speaker, we randomly crop a 3-second-long sample as the prompt and use the textual content of another 4 to 10-second-long sample as the input text. The inference procedure follows the methodology described in SoundStorm~\cite{borsos2023soundstorm}. As depicted in Figure~\ref{fig:coi_infer}, COI achieves the lowest WER and the highest speaker similarity across all iteration numbers compared to IG and SDG. This demonstrates the effectiveness of sequential semantic and perceptual information modeling in chain-of-information generation. Additionally, SDG's performance surpasses that of IG, indicating that as the degree of disentanglement increases, the effectiveness of speech generation also improves.

\begin{figure}[t]
    \centering
    \includegraphics[width=\linewidth]{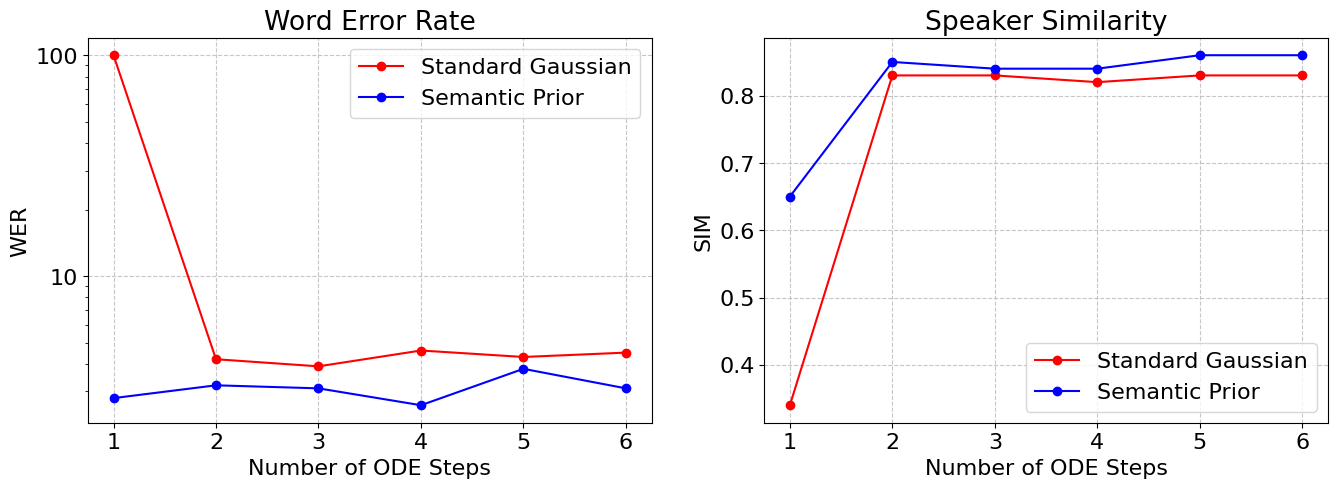}
    \caption{WER~(\textbf{Left}) and speaker similarity~(\textbf{Right}) of zero-shot voice conversion for flow matching with standard gaussian prior and semantic prior.}
    \label{fig:prior_analysis}
\vspace{-0.4cm}
\end{figure}

%% file: Sections/062_prior_analysis.tex
For perceptual modeling based on flow matching with an implicit chain, we modify the prior distribution from a standard Gaussian to a semantic Gaussian. We conduct experiments to assess whether the use of a semantic Gaussian can enhance the effectiveness of the flow matching process. We compare the performance of the flow matching model with a standard Gaussian prior distribution and a semantic Gaussian prior distribution in voice conversion. All other settings remain identical to those described in Section~\ref{sec:043_cnf}. We constructed a test set from VCTK and for each speaker, a 3-second-long sample was randomly cropped as the prompt, and another 4 to 10-second-long sample served as the source speech.
As illustrated in Figure~\ref{fig:prior_analysis}, when the ODE step is set to 1, the WER for the Standard Gaussian is 1, and the speaker similarity is very low. In contrast, the Semantic Gaussian exhibits an extremely low WER and decent speaker similarity, indicating that semantic Gaussian makes the inference process of flow matching more efficient. With the increase in ODE steps, the WER and speaker similarity corresponding to the Standard Gaussian consistently lag behind those of the Semantic Gaussian. This suggests that injecting semantic information into the prior for flow matching makes it more effective. We speculate that this is because flow matching learns the probability path from a simple prior distribution to a real data distribution, and semantic guassian prior is closer to the data distribution than standard guassian. Therefore, the probability path between them is easier to learn.

%% file: Sections/063_scaling_analysis.tex
\begin{figure}[t]
    \centering
    \includegraphics[width=\linewidth]{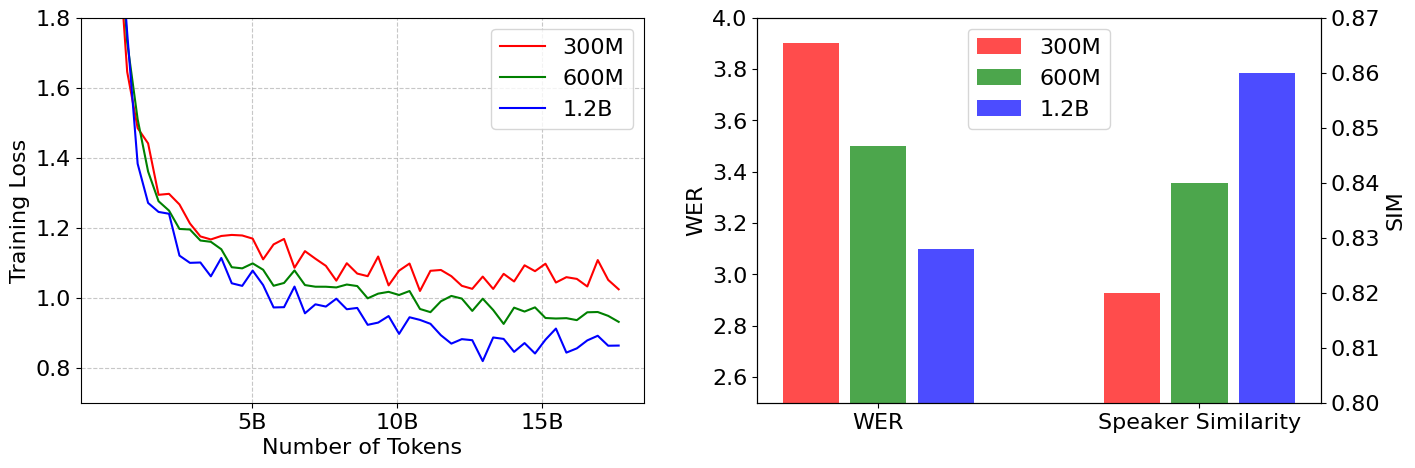}
    \caption{\textbf{Left:} Training los of flow matching models with different model sizes. \textbf{Right:} WER and speaker similarity of zero-shot voice conversion for flow matching with different model sizes.}
    \label{fig:scale}
\vspace{-0.4cm}
\end{figure}

Current large-scale speech generative models focus on data scaling, we explore the impact of model size scaling. We analyze the scaling behavior on training loss and performance on zero-shot voice conversion while increasing both model size and the number of tokens seen during training. 

\textbf{Training loss}~
As shown in Figure~\ref{fig:scale} on the left, with the increase in the number of tokens seen during training, the training loss consistently decreases for all models. Moreover, the loss decreases faster at larger model sizes. This demonstrates the strong scalability behavior of perceptual modeling models with larger model sizes and more data.

\textbf{Downstream Performance}~
We evaluate the performance of three model sizes on zero-shot voice conversion. The test set is constructed from the VCTK dataset. For each speaker, a 3-second-long sample was randomly cropped as the prompt, and another 4 to 10-second-long sample served as the source speech. The ode step is set to 8 for all models. As depicted in the right part of Figure~\ref{fig:scale}, with the increase in model size, WER consistently decreases, and Speaker Similarity continuously improves, demonstrating excellent scalability.

%% file: Sections/064_cont_dist_analysis.tex
\begin{figure}[t]
    \centering
    \includegraphics[width=0.8\linewidth]{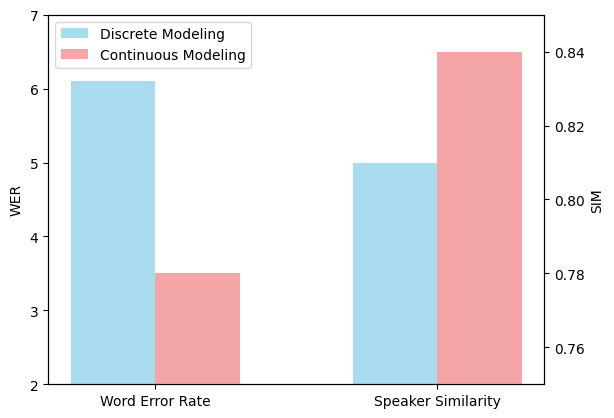}
    \caption{WER and speaker similarity of zero-shot voice conversion for discrete perceptual modeling and continuous perceptual modeling.}
    \label{fig:cont_vs_dist}
\vspace{-0.5cm}
\end{figure}

Flow matching based perceptual modeling utilizes continuous representation as input and output, modeling within a continuous space. On the other hand, discrete modeling based on speech representations is often employed in speech generation, with SoundStorm being a representative example. Therefore, we conducted experiments to compare which approach is more effective: continuous perceptual modeling or discrete perceptual modeling.
For continuous modeling, we selected Flow Matching with an implicit chain. For discrete modeling, we adopted the training and inference procedure of SoundStorm. Both models share a standard Conformer network architecture with bidirectional self-attention and rotary positional embeddings. Additionally, both models were trained on the Multilingual LibriSpeech dataset.
We present the results of zero-shot voice conversion using continuous and discrete modeling in Figure~\ref{fig:cont_vs_dist}. With both models having 8 inference iterations, continuous modeling outperforms discrete modeling in both WER and speaker similarity.

%% file: Sections/070_conclusion.tex
Addressing the redundancy in current SLLMs in modeling speech information, we introduce the Chain-of-Information Generation approach to decouple semantic and perceptual information. Furthermore, we trained an efficient, large-scale speech language model, Speech-Gen. 
By employing an autoregressive language model for semantic modeling and a non-autoregressive approach based on flow matching for perceptual modeling, we achieved superior text-to-speech generation in terms of quality and efficiency. 
Furthermore, we enhance the quality of speech synthesis by incorporating semantic information as a prior in flow matching. Speech-Gen demonstrates notably superior and more efficient performance in zero-shot text-to-speech, zero-shot voice conversion, and speech-to-speech dialogue tasks.

%% file: Sections/080_appendix.tex
\clearpage

\section{Model Configuration}
\label{sec:app:model_config}
\input{Sections/092_model_structure}

\section{Prompt for ChatGPT Score Evaluation}
\label{sec:app:prompt_chatgpt_score}
\input{Sections/091_prompt_chatgpt_score}

%% file: Sections/092_model_structure.tex
\begin{table}[htb]
    \small
    \centering
    \begin{tabular}{lcc|cccc}
        \toprule
         & \textbf{Model} & \textbf{Parameters} & \textbf{Layers} & \textbf{Hidden Dim} & \textbf{MLP Dim} & \textbf{Heads} 
          \\
        \midrule
        \textsc{Small} & \textbf{Conformer} & \textbf{300M} & 6 & 1024 & 4096 & 16  \\
        \textsc{Base} & \textbf{Conformer} & \textbf{600M} & 12 & 1024 & 4096 & 16  \\
        \textsc{Large} & \textbf{Conformer} & \textbf{1.2B} & 24 & 1024 & 4096 & 16  \\
        \bottomrule
    \end{tabular}
    \caption{Model configuration for different perceptual modeling models.}
    \label{tab:model_size}
\end{table}

\begin{table}[htb]
    \small
    \centering
    \begin{tabular}{lcc|cccc}
        \toprule
         & \textbf{Model} & \textbf{Parameters} & \textbf{Layers} & \textbf{Hidden Dim} & \textbf{MLP Dim} & \textbf{Heads} 
          \\
        \midrule
        \textsc{AR} & \textbf{Transformer} & \textbf{300M} & 12 & 1024 & 4096 & 16  \\
        \textsc{NAR} & \textbf{Conformer} & \textbf{600M} & 12 & 1024 & 4096 & 16  \\
        \bottomrule
    \end{tabular}
    \caption{Model configuration for autoregressive and non-autoregressive models for Section~\ref{sec:coi}.}
    \label{tab:model_size2}
\end{table}

%% file: Sections/091_prompt_chatgpt_score.tex
\begin{tcolorbox}[width=1\textwidth]
I will provide you with a scripts for the multiple characters' communication. Please evaluate and score the quality and logical coherence of the script content. Specific requirements are as follows: Please conduct a thorough examination of each dialogue's language quality, emotional expression, logical consistency, and overall reasonableness. Begin by evaluating the language of each dialogue, ensuring that it is natural, fluent, and free from grammatical and lexical errors. Pay attention to emotional expression to ensure that the dialogues adequately convey the characters' emotions. Emphasize the assessment of logical coherence and reasonableness in the dialogues, ensuring that the characters' speech and actions align with common sense and that their decisions and behaviors possess sufficient rationale within the plot development.\\
Below is the data:\\
$[$BEGIN DATA$]$\\
***\\
$[$scripts$]$: $\{scripts\}$\\
***\\
$[$Criterion$]$: content quality and logical coherence:\\
"1": "Poor - The script lacks clarity, with language that is unclear or inappropriate. Emotional expression is poorly conveyed, and the dialogue lacks logical coherence, making it difficult to follow or believe."

"2": "Below average - The script demonstrates some clarity, but language usage may be inconsistent or contain errors. Emotional expression is present but may be inconsistent or not well conveyed. The logical coherence of the dialogue is compromised at times, affecting believability."

"3": "Average - The script generally maintains clarity in language, with few errors or inconsistencies. Emotional expression is reasonably conveyed, and there is a moderate level of logical coherence in the dialogue. However, some aspects may still lack depth or believability."

"4": "Above average - The script is clear and well-written, with minimal language issues. Emotional expression is effectively conveyed, and the dialogue exhibits a high level of logical coherence. The interactions and decisions of the characters are mostly believable, contributing to the overall quality."

"5": "Excellent - The script is exceptionally well-crafted with clear, engaging language. Emotional expression is vivid and effectively communicates the characters' feelings. The dialogue demonstrates outstanding logical coherence, ensuring that the characters' actions and decisions align seamlessly with the plot. The overall content quality is exceptional."

$[$END DATA$]$

Does the scripts meet the criterion? My score is: $[$insert score based on the provided content quality and logical coherence criteria$]$.
\end{tcolorbox}